\documentclass[10pt,twocolumn,letterpaper]{article}

\usepackage[pagenumbers]{cvpr} 
\usepackage[T1]{fontenc}
\definecolor{cvprblue}{rgb}{0.21,0.49,0.74}
\usepackage[pagebackref,breaklinks,colorlinks,allcolors=cvprblue]{hyperref}

\usepackage{bm}
\usepackage{multirow}
\usepackage{array}
\usepackage{xcolor}
\definecolor{myyellow}{RGB}{255,215,0}
\usepackage{pifont}
\usepackage{tikz}
\usetikzlibrary{calc}
\usepackage{graphicx}
\graphicspath{{figs/}}
\def\confName{CVPR}
\def\confYear{2026}

\title{Semantic Audio-Visual Navigation in Continuous Environments}

\author{
    Yichen Zeng\textsuperscript{1,2}
    Hebaixu Wang\textsuperscript{1,2}
    Meng Liu\textsuperscript{2,3,*}
    Yu Zhou\textsuperscript{4,2}
    Chen Gao\textsuperscript{2,5,*}
    Kehan Chen\textsuperscript{6,7}
    Gongping Huang\textsuperscript{1,\thanks{Corresponding Author.}}
    \\
    \textsuperscript{1}Wuhan University  \quad
    \textsuperscript{2}Zhongguancun Academy \quad
    \textsuperscript{3}Shandong Jianzhu University
    \\
    \textsuperscript{4}Nankai University  \quad
    \textsuperscript{5}Tsinghua University \quad
    \textsuperscript{6}CASIA \quad
    \textsuperscript{7}UCAS
}

\begin{document}
\maketitle

\begin{abstract}
Audio-visual navigation enables embodied agents to navigate toward sound-emitting targets by leveraging both auditory and visual cues.
However, most existing approaches rely on precomputed room impulse responses (RIRs) for binaural audio rendering, restricting agents to discrete grid positions and leading to spatially discontinuous observations.
To establish a more realistic setting, we introduce Semantic Audio-Visual Navigation in Continuous Environments (SAVN-CE), where agents can move freely in 3D spaces and perceive temporally and spatially coherent audio-visual streams.
In this setting, targets may intermittently become silent or stop emitting sound entirely, causing agents to lose goal information.
To tackle this challenge, we propose MAGNet, a multimodal transformer-based model that jointly encodes spatial and semantic goal representations and integrates historical context with self-motion cues to enable memory-augmented goal reasoning.
Comprehensive experiments demonstrate that MAGNet significantly outperforms state-of-the-art methods, achieving up to a 12.1\% absolute improvement in success rate. These results also highlight its robustness to short-duration sounds and long-distance navigation scenarios.
The code is available at https://github.com/yichenzeng24/SAVN-CE.
\end{abstract}   
\section{Introduction}
\label{sec:intro}

Embodied navigation requires agents to autonomously reach targets in previously unseen environments using sensory inputs.
Most prior work has focused on either egocentric vision~\cite{zhu2017target, zhu2017visual, gupta2017cognitive, wu2019bayesian} or incorporating textual instructions as an additional modality~\cite{anderson2018vision, fried2018speaker, wang2019reinforced, chen2021history, chen2022think, hong2021vln, georgakis2022cross}.
However, visual perception is often insufficient in indoor environments, where targets may lie outside the agent's field of view or lack distinctive visual cues. 
For instance, an agent may need to respond to an elderly person falling in another room, locate a ringing phone in the house, or turn off a stove with boiling water.
In these situations, auditory perception provides critical complementary information, enabling agents to infer the locations and categories of otherwise invisible targets.

\begin{figure}[t]
    \centering
    \includegraphics[width=\linewidth]{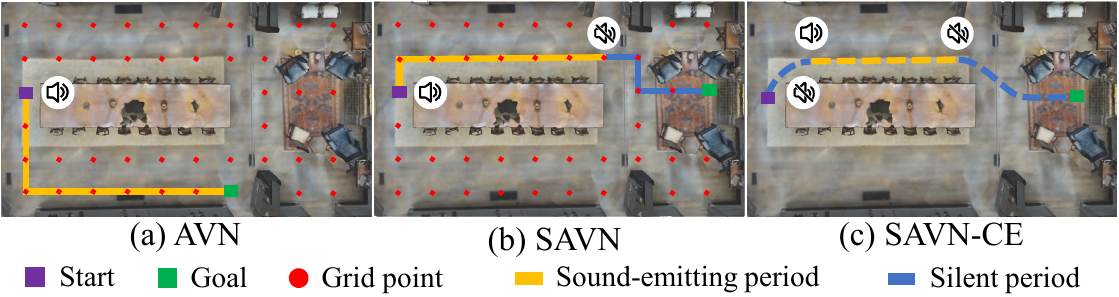}
    \caption{Illustration of the three navigation tasks: (a) the agent is restricted to discrete grid points and the sound-emitting goal is placed arbitrarily; (b) the goal, which emits a creaking sound for a limited duration, is semantically grounded in a chair; (c) the agent moves freely using fine-grained actions and the goal sound is available only within a short temporal window.}
    \label{fig:fig0}
    \vspace{-0.5em}
\end{figure}

Building on this motivation, audio-visual navigation (AVN)~\cite{gan2020look, chen2020soundspaces} enables agents to navigate toward sound-emitting goals in unmapped environments using audio-visual cues.
Agents do not have access to explicit goal information such as coordinates, object categories, or textual instructions.
Semantic audio-visual navigation (SAVN)~\cite{chen2021semantic} further extends this task by grounding short-duration audio signals in visual objects rather than arbitrary locations. 
However, both tasks rely on precomputed room impulse responses (RIRs), which demand terabytes of storage for binaural audio rendering.
This dependence further confines agents to discrete grid positions (1~m resolution) and four fixed orientations~\cite{chen2020soundspaces}, reducing task realism and hindering free exploration, as illustrated in \cref{fig:fig0}(a) and (b).

\begin{figure*}[t]
    \centering
    \includegraphics[width=0.96\linewidth]{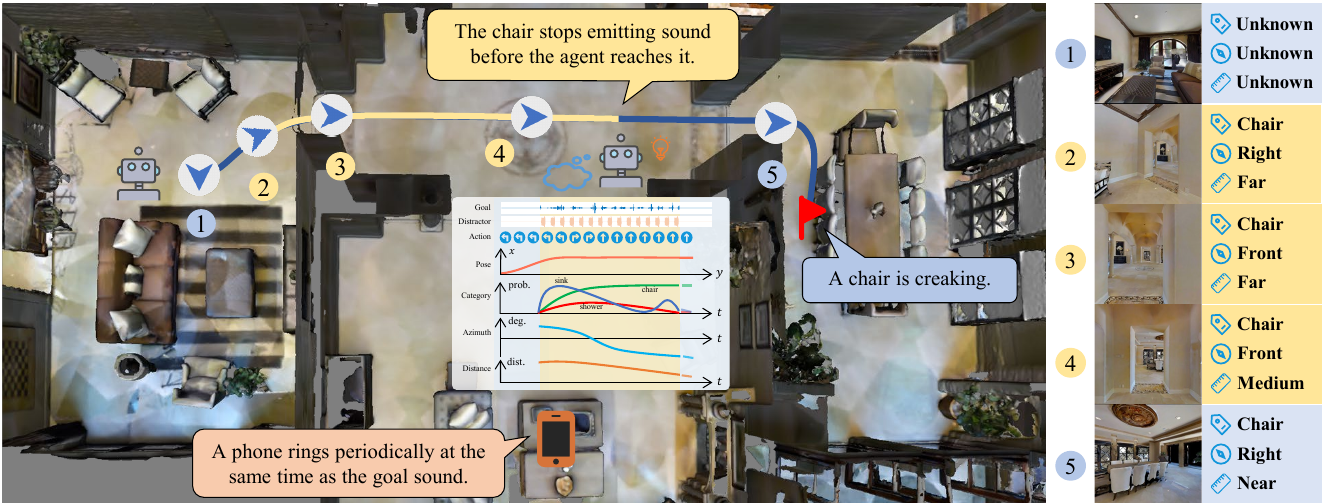}
    \caption{Overview of the proposed SAVN-CE framework. \ding{172} The agent is randomly initialized without prior knowledge of the environment or the goal. \ding{173} It explores the environment until the goal object (a chair) starts emitting sound. \ding{174} Leveraging multimodal cues, the agent infers the goal's semantic category, azimuth, and distance, and navigates toward it while avoiding obstacles and acoustic distractors (e.g., a ringing phone). As the agent approaches the goal (\ding{175} $\rightarrow$ \ding{176}), the sound emission ceases. Elements highlighted in \textcolor{myyellow}{yellow} and \textcolor{blue}{blue} denote sound-emitting and silent periods, respectively. \ding{176} During the silent period, the agent maintains goal tracking by integrating historical goal representations with current self-motion cues (e.g., the previous action and the current pose), successfully localizing and reaching the creaking chair despite the presence of visually similar objects and distractor sounds.}
    \label{fig:fig1}
    \vspace{-0.6em}
\end{figure*}

To bridge this gap, we introduce SAVN-CE (\emph{\textbf{S}emantic \textbf{A}udio-\textbf{V}isual \textbf{N}avigation in \textbf{C}ontinuous \textbf{E}nvironments}), a new task that allows agents to move freely in continuous 3D environments with fine-grained actions.
In this setting, agents can perceive temporally and spatially coherent observations, which enhances their ability to reason about how their position and orientation evolve relative to the goal as they move.
Unlike previous tasks, SAVN-CE features a highly dynamic goal condition, where the goal neither emits sound at the beginning nor persists until the end of an episode.
Consequently, agents must first explore the environment without any goal information and, once the goal begins to emit sound, execute long-horizon navigation toward it while avoiding obstacles, as illustrated in \cref{fig:fig0}(c).

The core challenge of SAVN-CE lies in accurately inferring both the spatial location and semantic category of the goal from partial sensory observations.
During the sound-emitting period, the goal may temporarily become silent due to large temporal gaps (e.g., intermittent creaking sounds) and a finer simulation step (0.25~s in our setting), making it difficult to continuously estimate its position and category.
Moreover, sound semantics are often ambiguous within short temporal windows and become distinguishable only when observed across longer temporal contexts.
The situation is further exacerbated once the goal stops emitting sound, resulting in a sustained loss of goal information.

Taking these challenges into consideration, we propose MAGNet (\emph{\textbf{M}emory-\textbf{A}ugmented \textbf{G}oal descriptor \textbf{Net}work}), a multimodal transformer-based architecture designed for reliable goal inference and efficient navigation.
MAGNet jointly encodes spatial and semantic goal representations and integrates historical context with self-motion cues for continuous goal inference.
By capturing temporal dependencies between past information and current sensory observations, MAGNet enables agents to navigate effectively toward the goal even after auditory signals are no longer available, as illustrated in \cref{fig:fig1}.

For a fair comparison, we adapt and retrain existing audio-visual navigation methods in continuous environments and systematically evaluate them on our SAVN-CE dataset, which is built upon Matterport3D~\cite{chang2017matterport3d}.
Experimental results show that MAGNet significantly outperforms prior methods, particularly in challenging scenarios involving short-duration sounds and long-distance navigation.

In summary, our contributions are threefold:
\begin{itemize}
    \item We introduce SAVN-CE, which extends semantic audio-visual navigation to continuous environments, bringing the task closer to realistic scenarios.
    \item We propose MAGNet, which leverages historical context and self-motion cues for robust goal reasoning and efficient navigation, thereby addressing the challenge of goal information loss when the goal sound is absent.
    \item Extensive experiments demonstrate that MAGNet significantly outperforms existing methods, achieving up to a 12.1\% absolute improvement in success rate.
\end{itemize}
\section{Related Work}
\label{sec:related_work}

\noindent\textbf{Sound event localization and detection.}
\label{sec:seld}
Sound event localization and detection (SELD)~\cite{adavanne2019sound} unifies two fundamental auditory perception tasks: sound event detection~\cite{cakir2017convolutional, xia2020sound} and sound source localization~\cite{adavanne2018direction, perotin2019crnn, grumiaux2022survey}, thereby providing a comprehensive spatiotemporal representation of acoustic scenes.
It jointly estimates the temporal boundaries, categories, and spatial positions of sound events within a unified vector framework~\cite{shimada2021accdoa, shimada2022multi}.
Recent advances have extended SELD to handle more challenging scenarios, including multiple and moving sound sources~\cite{shimada2022multi}, and have also incorporated visual cues~\cite{berghi2024fusion} to further enhance performance.
However, most existing approaches rely on either simulated datasets~\cite{adavanne2019sound, hu2025pseldnets} or single-room recordings~\cite{shimada2023starss23}, leaving realistic, acoustically complex multi-room environments largely underexplored.

\noindent\textbf{Semantic audio-visual navigation.}
\label{sec:savn}
Although subsequent studies have extended AVN to more complex settings, such as sound-attacker~\cite{yu2022sound}, multi-goal~\cite{kondoh2023multi}, and moving-source scenarios~\cite{younes2023catch}, the standard formulation still exhibits several key limitations: 1) the goal emits sound continuously throughout the episode; 2) the goal's position is placed arbitrarily in the scene, lacking any visual embodiment; and 3) agents are confined to predefined grid locations where precomputed RIRs are available.
To mitigate the first two issues, SAVN~\cite{chen2021semantic} was introduced to enable agents to navigate toward semantically grounded sound-emitting objects.
Subsequent efforts have further enriched this framework by incorporating language instructions~\cite{paul2022avlen, liu2024caven}, leveraging large language models~\cite{yang2024rila}, or handling multiple sound sources~\cite{shi2025towards}.
However, the third limitation remains largely unsolved~\cite{kondoh2023multi, chen2024sim2real}, and no prior work has explored it within the SAVN paradigm, despite the capability of SoundSpaces 2.0~\cite{chen2022soundspaces} to support continuous navigation in realistic 3D environments.
The principal difficulty lies in the complexity of the training procedure, which is further aggravated by the limited simulation speed resulting from the high computational cost of binaural audio rendering.

\noindent\textbf{VLN in continuous environments.}
\label{sec:vlnce}
Vision-and-language navigation in continuous environments (VLN-CE) was introduced by Krantz et al.~\cite{krantz2020beyond} to eliminate the discrete graph assumptions of the original VLN setting~\cite{anderson2018vision}.
In VLN-CE, agents must execute low-level actions within 3D continuous environments to reach a goal by following language instructions and visual observations, without access to global topology, oracle navigation, or perfect localization.
Consequently, these requirements make the task substantially more challenging than discrete VLN, resulting in much lower absolute performance.
To address these challenges, recent research has focused on waypoint prediction for long-range planning~\cite{krantz2021waypoint, hong2022bridging, krantz2022sim, an2024etpnav, chen2025constraint} and obstacle-avoidance strategies to prevent agents from getting stuck~\cite{an2024etpnav, yue2024safe}.
In contrast, SAVN-CE centers on inferring goal information from partial sensory observations even after the goal sound has ceased.
\section{SAVN in Continuous Environments}
\label{sec:SAVNCE}

We propose SAVN-CE, a new task that requires embodied agents to navigate toward semantic sound-emitting goals in continuous 3D indoor environments using fine-grained, low-level actions.
Unlike prior settings that rely on precomputed RIRs, which consume terabytes of storage, SAVN-CE allows agents to move freely in continuous spaces with temporally and spatially coherent audio rendered dynamically.
This formulation demands that agents effectively integrate multimodal sensory inputs and perform long-horizon reasoning to reach the goal.

\noindent\textbf{Simulator.}
\label{sec:simulator}
SAVN-CE is implemented on SoundSpaces 2.0~\cite{chen2022soundspaces}, an extension of Habitat~\cite{savva2019habitat, szot2021habitat} that supports continuous audio rendering within realistic Matterport3D scenes~\cite{chang2017matterport3d}.
The simulator operates at a 16~kHz audio sampling rate with a 0.25~s simulation step, corresponding to 4,000 audio samples per step.
Since the reverberation time of binaural RIRs rendered in Matterport3D scenes is typically much longer than a single simulation step, considering only the RIRs of the current and previous steps is insufficient to model long-tail reverberation effects (as done in SoundSpaces 2.0).
To address this issue, following~\cite{scheibler2018pyroomacoustics, diaz2021gpurir}, we convolve the source sound with the current-step binaural RIRs and accumulate the residual responses from all previous steps to generate temporally coherent audio.

\noindent\textbf{Actions and observations.}
\label{sec:actions_and_observations}
The agent's action space comprises four discrete actions: \emph{MoveForward 0.25~m}, \emph{TurnLeft 15${^\circ}$}, \emph{TurnRight 15${^\circ}$}, and \emph{Stop}, consistent with VLN-CE~\cite{krantz2020beyond}.
Observations include binaural audio waveforms (mimicking human spatial hearing), egocentric RGB-D images (128$\times$128 pixels, 90$^\circ$ field-of-view), and the agent's pose relative to its initial position and orientation.

\noindent\textbf{Dataset construction.}
\label{sec:dataset_construction}
Each episode is defined by: 
1) the scene, 
2) the agent's initial location and orientation, 
3) the goal's location and semantic category, and
4) the onset time and duration of the goal sound.
When a distractor is included, its location and category are also specified, and it shares identical temporal boundaries with the goal sound. 
The onset time of the goal sound is uniformly sampled from $[0, 5]$~s, while its duration follows a Gaussian distribution with a mean of 15~s and a standard deviation of 9~s. 
We adapt the dataset from SAVi~\cite{chen2021semantic} to construct our SAVN-CE dataset, adopting the same 21 semantic categories as goal objects and 102 periodic sounds from SoundSpaces~\cite{chen2020soundspaces} as distractor candidates, disjoint from the goal categories. 
This setup ensures temporal diversity and acoustic ambiguity, making the task more challenging and realistic.
Our dataset contains 0.5M/500/1,000 episodes for train/val/test, respectively.
These splits use disjoint sets of scenes and source sounds, requiring agents to generalize to both unseen environments and unheard sounds. 
In the test split, the average number of oracle actions is 78.49, substantially higher than 26.52 in the discrete setting.

\noindent\textbf{Success criterion.}
\label{sec:success_criterion}
An episode is considered successful if the agent issues the \emph{Stop} action within 1~m of the target sound source.
Stopping near the distractor or another instance of the same semantic category is regarded as a failure.
Following prior work in audio-visual navigation~\cite{chen2020soundspaces, chen2021semantic}, each episode is limited to at most 500 actions.  

\section{Method}
\label{sec:method}

\begin{figure*}[t]
    \centering
    \begin{tikzpicture}
    \node[anchor=south west,inner sep=0] (image) at (0,0) {\includegraphics[width=0.90\linewidth]{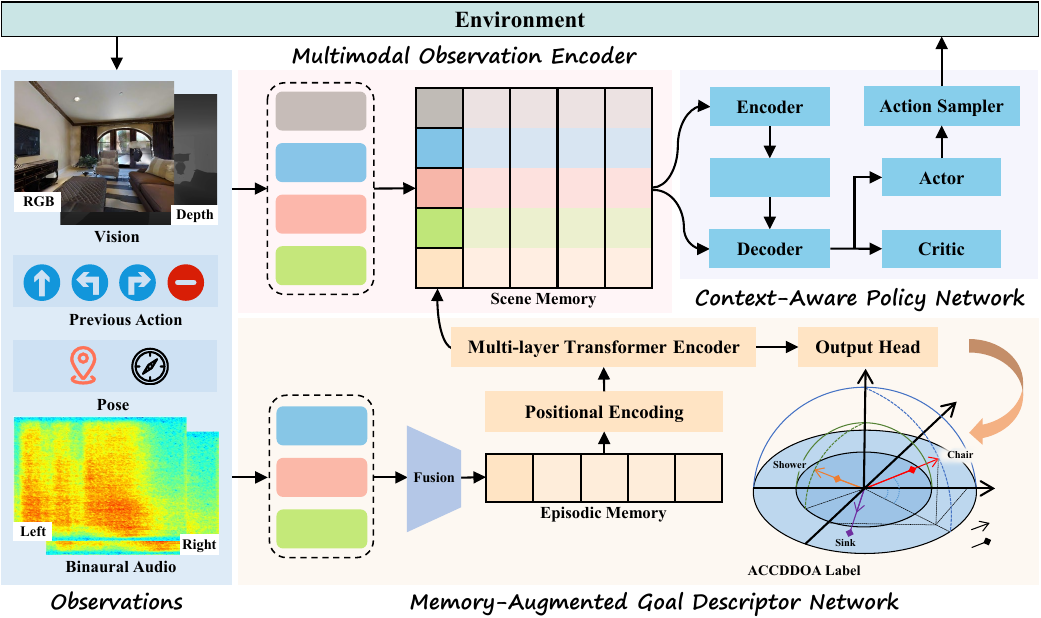}};
    \begin{scope}[x={(image.south east)},y={(image.north west)}]
        \node at (0.140, 0.908) {\normalsize $\bm{O}_{t}$};
        \node at (0.926, 0.906) {\normalsize $a_{t}$};
        \coordinate (A) at (0.309, 0.569);
        \coordinate (B) at (0.423, 0.566);
        \foreach \i/\txt in {0/{Audio Encoder}, 1/{Pose Encoder}, 2/{Action Encoder}, 3/{Visual Encoder}}{
            \node at ($(A) + (0, \i*0.083)$) {\fontsize{6pt}{7pt}\selectfont \txt};
        }
        \foreach \i/\txt in {0/$\bm{e}^{G}_{t}$, 1/$\bm{e}^{B}_{t}$, 2/$\bm{e}^{p}_{t}$, 3/$\bm{e}^{a}_{t}$, 4/$\bm{e}^{I}_{t}$}{
            \node at ($(B) + (0, \i*0.065)$) {\scriptsize \txt};
        }
        \coordinate (C) at (0.309, 0.146);
        \foreach \i/\txt in {0/{Audio Encoder}, 1/{Pose Encoder}, 2/{Action Encoder}}{
            \node at ($(C) + (0, \i*0.082)$) {\fontsize{6pt}{7pt}\selectfont \txt};
        }
        \node at (0.514, 0.690) {\footnotesize $\cdots$};
        \node at (0.604, 0.690) {\footnotesize $\bm{e}^{O}_{t^{\prime}}$};
        \node at (0.490, 0.224) {\footnotesize $\bm{m}_{t}$};
        \node at (0.582, 0.224) {\footnotesize $\cdots$};
        \node at (0.672, 0.224) {\footnotesize $\bm{m}_{t^{\prime}}$};
        \node at (0.741, 0.712) {\footnotesize $\bm{M}_{e}$};
        \node at (0.964, 0.210) {\small $x$};
        \node at (0.928, 0.353) {\small $y$};
        \node at (0.846, 0.390) {\small $z$};
        \node at (0.850, 0.136) {\tiny $x_{ct}$};
        \node at (0.925, 0.191) {\tiny $y_{ct}$};
        \node at (0.900, 0.178) {\tiny $z_{ct}$};
        \node at (0.857, 0.199) {\tiny $\phi$};
        \node at (0.868, 0.198) {\tiny $\theta$};
        \node at (0.970, 0.120) {\scriptsize $d_{ct}$};
        \node at (0.972, 0.152) {\scriptsize $\bm{R}_{ct}$};
        \node at (0.882, 0.078) {\scriptsize $\left[a_{ct}\bm{R}_{ct}, d_{ct}\right]$};
    \end{scope}
    \end{tikzpicture}
    \caption{
        Overall architecture of MAGNet.
        The \emph{multimodal observation encoder} extracts multimodal features from current sensory inputs and updates the scene memory accordingly.
        The \emph{memory-augmented goal descriptor network} infers spatial and semantic representations of the goal by integrating auditory cues, self-motion cues, and historical goal embeddings stored in the episodic memory, thereby ensuring temporally consistent inference even after the goal sound ceases entirely.
        Conditioned on the latest scene memory embeddings, the \emph{context-aware policy network} attends to the encoded memory $\bm{M}_{e}$ to predict the next action, enabling continuous navigation toward the goal.
    }
    \label{fig:model_architecture}
\end{figure*}
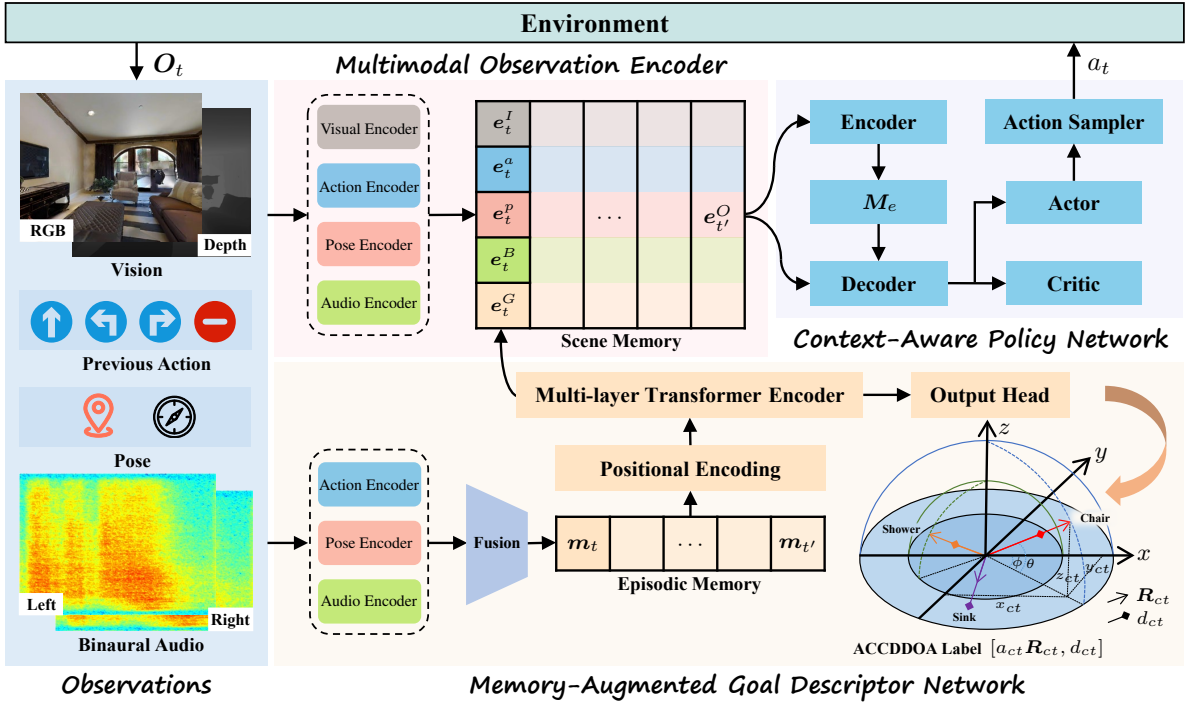

To tackle the challenges of SAVN-CE, particularly in maintaining goal awareness when the goal sound becomes intermittent or silent, we propose MAGNet, a multimodal transformer-based architecture to enable robust goal reasoning and efficient navigation. 
As illustrated in~\cref{fig:model_architecture}, MAGNet consists of three modules: 
1) Multimodal Observation Encoder, which transforms multimodal inputs into compact embeddings and stores them in a long-term scene memory~\cite{fang2019scene, chen2021semantic}; 
2) Memory-Augmented Goal Descriptor Network, which fuses auditory cues, egocentric motion information, and episodic memory to maintain a stable goal representation, ensuring persistent tracking even after the sound ceases; and 
3) Context-Aware Policy Network, which attends to the aggregated scene memory to predict the next action of the agent. 

\subsection{Multimodal Observation Encoder}
\label{sec:observation_encoder}

At time step $t$, the agent receives multimodal observations $\bm{O}_{t} \! = \! \{\bm{I}_{t}, a_{t-1}, \bm{p}_{t}, \bm{B}_{t}\}$, where $\bm{I}_{t}$ denotes the RGB-D images, $a_{t-1}$ indicates the action taken at the previous time step, $\bm{p}_{t} \! = \! [x_{t}, y_{t}, \theta_{t}, t]$ is the agent's current pose, and $\bm{B}_{t}$ represents the binaural audio input.
The observation encoder module consists of four modality-specific encoders (visual, action, pose, and audio), which process their corresponding inputs into embeddings.
The concatenation of these embeddings constitutes the observation representation: 
$\bm{e}^{O}_{t} \! = \! \left[ \bm{e}^{I}_{t}, \bm{e}^{a}_{t}, \bm{e}^{p}_{t}, \bm{e}^{B}_{t}, \bm{e}^{G}_{t} \right]$, where $\bm{e}^{G}_{t}$ is the goal embedding described in \cref{sec:goal_descriptor_network}.
The scene memory maintains the most recent $N_{s}$ encoded observations:
\begin{equation}
    \bm{M}_{s,t} = \left\{ \bm{e}_{t^{\prime}}^{O} \,\middle|\, t^{\prime} \in \left[ \max\left(0,\, t - N_{s} + 1\right), t \, \right] \right\}.
    \label{eq:scene_memory}
\end{equation}

The visual encoder processes RGB and depth images using two independent ResNet-18 backbones~\cite{he2016deep}, producing a concatenated embedding $\bm{e}^{I}_{t}$.
The previous action is mapped into the action embedding $\bm{e}^{a}_{t}$ via an embedding layer.
The agent's pose $\bm{p}_{t}$ is normalized to $\left[x_{t}/d, y_{t}/d, \sin(\theta_{t}), \cos(\theta_{t}), t/t_{\text{max}}\right]$, where $d$ is a distance scale factor and $t_{\text{max}}$ denotes the maximum episode length. The normalized pose is projected into the pose embedding $\bm{e}^{p}_{t}$ through a fully connected layer.

The binaural waveforms are transformed into complex spectrograms using a short-time Fourier transform (STFT) with a 512-point FFT and a hop length of 160 samples.
To jointly encode spatial and semantic acoustic cues, we compute four complementary channels: the mean magnitude spectrogram, the sine and cosine components of the inter-channel phase difference, and the inter-channel level difference~\cite{krause2023binaural, shi2025towards}.
The audio embedding $\bm{e}^{B}_{t}$ is extracted from these features by the audio encoder, which comprises three convolutional layers followed by a fully connected layer.
See Supp. for details of the acoustic feature extraction.

\subsection{Memory-Augmented Goal Descriptor Network}
\label{sec:goal_descriptor_network}

When the goal sound is intermittent or completely silent, maintaining a stable goal representation is essential for reliable navigation.
To this end, we propose a memory-augmented goal descriptor network (GDN) that fuses binaural features, self-motion cues, and episodic memory to model temporal continuity and spatial dynamics explicitly.
Unlike prior approaches that rely solely on current binaural audio, our design captures the evolving spatial relationship between the agent and the goal over time by combining auditory and self-motion cues, while leveraging episodic memory to maintain temporal continuity.

Self-motion cues, including the agent's previous action and current pose, are crucial for estimating how the goal's relative position changes as the agent moves.
Specifically, the \emph{TurnLeft} and \emph{TurnRight} actions decrease and increase the goal's azimuth relative to the agent by 15$^\circ$, respectively, while the \emph{MoveForward} action affects both azimuth and distance depending on the agent's current position.
The agent's pose $\bm{p}_{t}$, encoding its translation and orientation relative to the initial state, further enhances spatial reasoning in binaural sound source localization~\cite{krause2023binaural, garcia2022binaural}.

Since real-world acoustic events are often intermittent or ambiguous, the network must reason over historical auditory context rather than isolated observations.
To capture such temporal dynamics, we adopt an episodic memory module~\cite{pashevich2021episodic}, which stores goal-relevant embeddings from past steps.
At time step $t$, the GDN receives the binaural audio $\bm{B}_{t}$, the previous action $a_{t-1}$, and the current pose $\bm{p}_{t}$. These encoders follow the same structures as in~\cref{sec:observation_encoder}, except that the audio encoder produces a higher-dimensional embedding to better capture spatial and semantic goal information.
The three embeddings are fused through a multi-layer perceptron (MLP) into a unified representation: $\bm{m}_{t} = \text{MLP}\left[ \bm{e}_{t}^{a} ,\, \bm{e}_{t}^{p} ,\, \bm{e}_{t}^{B} \right]$, which is then appended to the episodic memory:
\begin{equation}
    \bm{M}_{g,t} = \left\{ \bm{m}_{t^{\prime}} \,\middle|\, t^{\prime} \in \left[ \max\left(0,\, t - N_{g} + 1\right), t \, \right] \right\},
    \label{eq:episodic_memory}
\end{equation}
where $N_{g}$ is the episodic memory capacity.
The collected episodic memory is augmented with positional encodings to preserve temporal order and then processed by a transformer encoder with two output branches.

The first branch projects the encoder output through a fully connected layer to obtain the goal embedding: $\bm{e}^{G}_{t} = \text{FC} \left( \left[ \text{Encoder}(\bm{M}_{g,t}) \right]_{t} \right)$,
where the subscript $t$ denotes the encoder output corresponding to the current time step.
The second branch, in contrast, outputs goal descriptions in the activity-coupled Cartesian distance and direction-of-arrival (ACCDDOA) format~\cite{shimada2021accdoa, krause2024sound} using an MLP output head, which is employed for loss computation and network optimization during training.
As illustrated in the bottom-right corner of \cref{fig:model_architecture}, the ACCDDOA-formatted goal descriptions are formulated as: $\bm{y}_{ct} \! = \! [a_{ct}\bm{R}_{ct}, d_{ct}]$, where $c$ and $t$ denote the category and time step indices, respectively.
Here, $\bm{R}_{ct} \! = \! \left[x_{ct}, y_{ct}, z_{ct}\right] \! \in \! [-1, 1]^{3}$ represents the unit-norm direction-of-arrival (DOA) vector, $a_{ct} \! \in \! \{0, 1\}$ indicates the sound activity status (0 for inactive and 1 for active), and $d_{ct}\!>\!0$ is the normalized distance.

By integrating self-motion dynamics and temporally accumulated goal embeddings, the memory-augmented GDN preserves consistent goal representations even in the absence of auditory input.
The fine-grained action space further limits the positional changes between consecutive steps, ensuring stable and coherent goal tracking throughout long-horizon navigation in continuous environments.

\subsection{Context-Aware Policy Network}
\label{sec:policy_network}

The context-aware policy network employs a transformer-based encoder-decoder architecture to facilitate temporally informed decision-making by integrating both historical and current observations.
At each time step $t$ during an episode, the encoder processes the accumulated scene memory $\bm{M}_{s,t}$ to capture temporal dependencies across past observations, yielding an encoded representation $\bm{M}_{e} \! = \! \text{Encoder}(\bm{M}_{s,t})$.
The decoder then generates a context-aware latent state representation:
\begin{equation}
    \bm{s}_{t} = \text{Decoder}(\bm{M}_{e}, \text{MLP}\left( \bm{e}^{O}_{t} \right)),
\end{equation}
which serves as a compact summary of both historical and current sensory information.
This representation is passed separately to an actor and a critic, each implemented as a fully connected layer that predicts the action distribution and the state value, respectively.
Finally, an action sampler selects the next action $a_{t}$ from the predicted distribution, enabling the agent to execute coherent and contextually grounded actions throughout the episode in a partially observable continuous environment.

\subsection{Training Strategy}
\label{sec:training_strategy}

For stable and efficient training, each iteration consists of a 150-step rollout with the current policy network, followed by updates to both the GDN and the policy network using the collected experiences. 

The GDN is trained online in a supervised manner using complete episodes.
To construct these episodes, unfinished episodes from the previous iteration are merged with newly collected ones, ensuring full temporal continuity. 
Episodes shorter than 30 steps are discarded to guarantee that the goal has emitted sound by the end of the episode.
The training procedure leverages oracle ACCDDOA labels with a mean squared error (MSE) loss and the Adam~\cite{kingma2015adam} optimizer at a learning rate of $1 \times 10^{-3}$.
To maintain temporal causality, the encoder employs causal attention, preventing information leakage from future time steps.

The policy network is trained with decentralized distributed proximal policy optimization (DD-PPO)~\cite{wijmans2020ddppo}, following the two-stage paradigm used in SAVi~\cite{chen2021semantic}. It is optimized with the standard PPO loss~\cite{schulman2017proximal} using the Adam~\cite{kingma2015adam} optimizer with a learning rate of 2.5 $\times$ 10$^{-4}$.
The reward function has three components: a success reward of +10 for reaching the goal, an intermediate reward proportional to the change in geodesic distance to the goal, and a small time penalty of -0.01 per step to encourage efficient navigation.
\section{Experiments}
\label{sec:experiments}

\subsection{Experimental Setup}
\label{sec:experimental_setup}

\noindent\textbf{Baselines.}
We use the following methods for comparison: 

\begin{itemize}
    \item \textbf{Random}: A non-learning policy that samples actions according to the action distribution in the train split~\cite{krantz2020beyond}. 
    
    \item \textbf{ObjectGoal}: A policy where the agent receives RGB-D observations and the ground-truth goal category, without access to a perfect \emph{Stop} action.

    \item \textbf{AV-Nav}: A policy that leverages audio-visual input and employs a GRU to encode past observations~\cite{chen2020soundspaces}.

    \item \textbf{SMT + Audio}: A policy that integrates audio with other sensory observations through a scene memory transformer (SMT)~\cite{fang2019scene}, without explicit goal inference.

    \item \textbf{SAVi}: A policy designed for the SAVN task, employing a GDN to independently infer the goal's position and category from audio cues, and aggregating estimates over time using a weighting factor $\lambda$~\cite{chen2021semantic}.

    \item \textbf{Oracle}: A variant of our architecture without the audio encoder or GDN, but with access to oracle ACCDDOA labels. Oracle1 has access only during the sound-emitting period, whereas Oracle2 retains access until the episode ends, providing upper-bound performance when goal information is lost or maintained after sound ceases.
\end{itemize}

\noindent\textbf{Evaluation metrics.} 
Navigation performance is evaluated using the following metrics: 
1) \textbf{SR}: the fraction of successful episodes; 
2) \textbf{SPL}: success weighted by the ratio of the shortest path length to the agent's trajectory length~\cite{anderson2018evaluation}; 
3) \textbf{SNA}: success weighted by the ratio of the oracle number of actions to the agent's executed actions, which penalizes collisions and unnecessary rotations~\cite{chen2021learning}; 
4) \textbf{DTG}: the average geodesic distance to the goal at the end of the episode; 
5) \textbf{SWS}: the fraction of successful episodes in which the agent reaches the goal while it is silent~\cite{chen2021semantic}.
To assess the agent's susceptibility to distractors, we introduce a diagnostic metric, \textbf{DSR} (Distractor Success Rate), which measures the fraction of episodes in which the agent reaches the distractor incorrectly.
DTG is measured in meters, whereas all other metrics are reported as percentages.

\noindent\textbf{Implementation details.}
All learning-based approaches are trained for up to 240M steps, with early stopping applied when no improvement is observed on the validation split for 24M consecutive steps.
Training details follow those described in \cref{sec:training_strategy}.
The maximum capacities of the scene and episodic memories are set to $N_{s} \! = \! 150$ and $N_{g} \! = \! 128$, respectively.
The scale factor $d$ is set to 20~m to normalize the agent's pose and goal positions.
For models using binaural audio, training the full 240M steps takes roughly 14 days on 128 CPU threads and 4 NVIDIA A800 GPUs.

\begin{table*}[!htbp]
    \centering
    \normalsize
    \belowrulesep=0pt
    \aboverulesep=0pt
    \caption{Navigation performance on the SAVN-CE dataset under \emph{Clean} and \emph{Distracted Environments}. }
    \resizebox{0.95\textwidth}{!}{
    \begin{tabular}{c|rrrrr|rrrrrr}
    \toprule
    & \multicolumn{5}{c|}{\emph{Clean Environments}} & \multicolumn{6}{c}{\emph{Distracted Environments}} \\
    \cmidrule(lr){2-6} \cmidrule(lr){7-12}
    & SR\raisebox{0.25ex}{$\uparrow$} & SPL\raisebox{0.25ex}{$\uparrow$} & SNA\raisebox{0.25ex}{$\uparrow$}
    & DTG\raisebox{0.25ex}{$\downarrow$} & \makebox[1.5em][c]{SWS\raisebox{0.25ex}{$\uparrow$}}

    & SR\raisebox{0.25ex}{$\uparrow$} & SPL\raisebox{0.25ex}{$\uparrow$} & SNA\raisebox{0.25ex}{$\uparrow$}
    & DTG\raisebox{0.25ex}{$\downarrow$} & SWS\raisebox{0.25ex}{$\uparrow$} & DSR \\

    \midrule
    Random              & 0.3   & 0.2\hspace{0.3em}     & 0.1\hspace{0.4em}     & 15.7\hspace{0.5em}    & 0.2\hspace{0.3em} 
                        & 0.4   & 0.3\hspace{0.3em}     & 0.1\hspace{0.4em}     & 15.8\hspace{0.5em}    & 0.4\hspace{0.3em} & 1.2\hspace{0.3em} \\

    ObjectGoal          & 0.8   & 0.6\hspace{0.3em}     & 0.2\hspace{0.4em}     & 14.4\hspace{0.5em}    & 0.4\hspace{0.3em} 
                        & 1.1   & 0.8\hspace{0.3em}     & 0.4\hspace{0.4em}     & 14.4\hspace{0.5em}    & 0.8\hspace{0.3em} & 0.8\hspace{0.3em} \\

    AV-Nav              & 21.3  & 17.8\hspace{0.3em}    & 13.1\hspace{0.4em}    & 10.7\hspace{0.5em}    & 4.0\hspace{0.3em} 
                        & 13.5  & 10.9\hspace{0.3em}    & 7.6\hspace{0.4em}     & 12.5\hspace{0.5em}    & 3.0\hspace{0.3em} & 6.7\hspace{0.3em} \\

    SMT + Audio         & 24.8  & 21.0\hspace{0.3em}    & 16.8\hspace{0.4em}    & 10.1\hspace{0.5em}    & 5.3\hspace{0.3em}  
                        & 16.5  & 13.7\hspace{0.3em}    & 9.4\hspace{0.4em}     & 11.8\hspace{0.5em}    & 5.0\hspace{0.3em} & 6.3\hspace{0.3em} \\

    SAVi                & 25.6  & 21.2\hspace{0.3em}    & 17.3\hspace{0.4em}    & 10.1\hspace{0.5em}    & 6.0\hspace{0.3em} 
                        & 18.5  & 14.9\hspace{0.3em}    & 11.1\hspace{0.4em}    
                        & \textbf{10.4}\hspace{0.5em}    & \textbf{5.4}\hspace{0.3em} & 6.4\hspace{0.3em} \\

    MAGNet (Ours)       & \textbf{37.7} & \textbf{32.9}\hspace{0.3em} & \textbf{27.4}\hspace{0.4em} 
                        & \textbf{8.0}\hspace{0.5em} & \textbf{10.6}\hspace{0.3em} 
                        & \textbf{19.3} & \textbf{16.5}\hspace{0.3em} & \textbf{13.4}\hspace{0.4em} 
                        & 10.6\hspace{0.5em} & 4.8\hspace{0.3em} & 7.8\hspace{0.3em} \\
    \midrule              
    Oracle1             & 41.4  & 37.8\hspace{0.3em}    & 31.0\hspace{0.4em}    & 6.3\hspace{0.5em}     & 13.0\hspace{0.3em} 
                        & 41.5  & 38.0\hspace{0.3em}    & 31.0\hspace{0.4em}    & 6.3\hspace{0.5em}     & 11.2\hspace{0.3em} & 0.5\hspace{0.3em} \\
    Oracle2             & 75.0     & 63.7\hspace{0.3em}        & 51.9\hspace{0.4em}        & 4.2\hspace{0.5em}        & 48.4\hspace{0.3em}  
                        & 74.3     &  64.0\hspace{0.3em}       & 51.8\hspace{0.4em}        & 4.5\hspace{0.5em}        & 45.4\hspace{0.3em} & 0.8\hspace{0.3em} \\
    \bottomrule
    \end{tabular}}

    \label{tab:clean_distracted}
\end{table*}

\begin{figure*}[t]
    \centering
    \includegraphics[width=\linewidth]{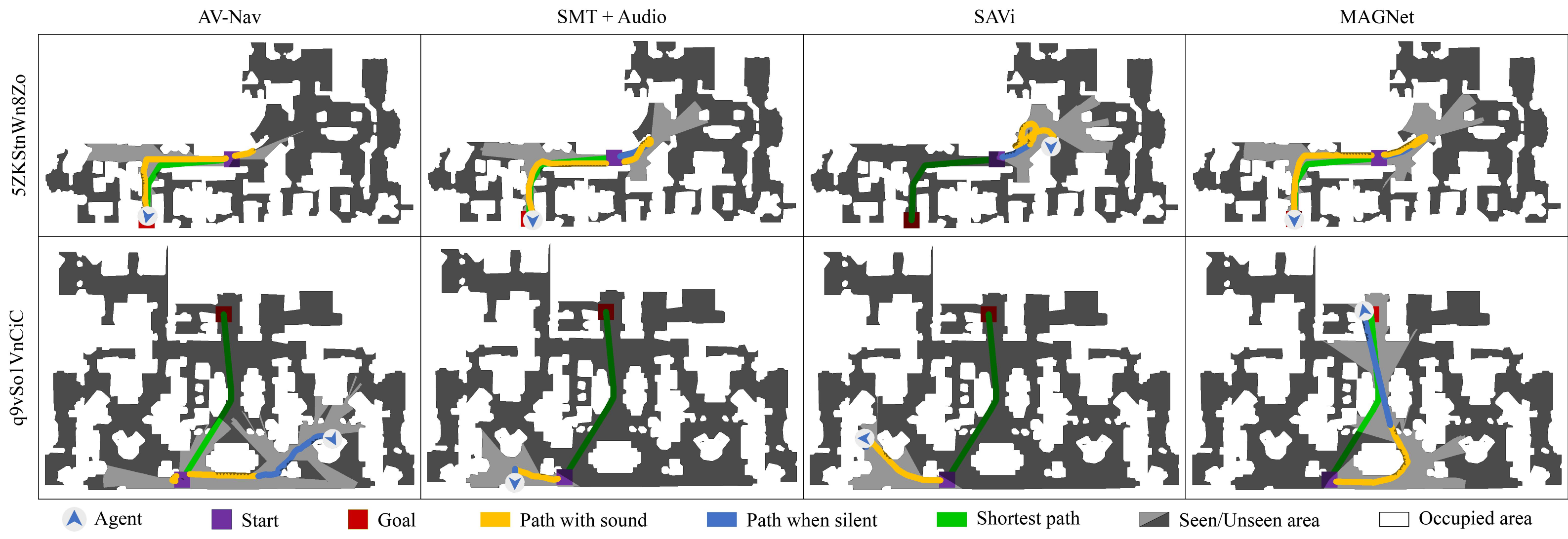}
    \caption{Navigation trajectories of different methods under \emph{Clean Environments}.}
    \label{fig:trajectories_clean}
\end{figure*}

\subsection{Experimental Results}
\label{sec:experimental_results}

We consider two experimental scenarios: 
1) \emph{Clean Environments}, where only the goal emits sound; and 
2) \emph{Distracted Environments}, where an additional distractor emits sound simultaneously.
Following prior work, we select the checkpoint achieving the highest SPL on the validation split for evaluation.
\cref{tab:clean_distracted} summarizes the navigation performance averaged over 1,000 test episodes and five independent runs.
Overall, MAGNet consistently outperforms all baselines by a large margin across all metrics under \emph{Clean Environments}, while the performance gain decreases under \emph{Distracted Environments} due to interference from distractor sounds.

\noindent\textbf{Performance under Clean Environments.}
Consistent with intuition, Random performs poorly, while ObjectGoal yields only marginal improvement, indicating that merely knowing the goal category without access to audio cues is insufficient for successful navigation.
AV-Nav and SMT + Audio show notable improvements but remain limited, reflecting their constrained capacity to exploit audio cues.
Among all baselines, SAVi achieves the best performance by explicitly inferring the goal's position and category through its GDN.
Nevertheless, MAGNet surpasses SAVi by a substantial margin across all metrics, highlighting the effectiveness of our memory-augmented GDN and multimodal fusion strategy in enabling reliable goal reasoning and efficient navigation (see SWS and SPL metrics).

\noindent\textbf{Performance under Distracted Environments.}
All methods exhibit degraded performance compared to \emph{Clean Environments}, indicating that distractor sounds significantly hinder agents from focusing on goals. Although an additional ground-truth goal category is provided to help agents distinguish the goal from the distractor, they still struggle to locate it accurately.
In particular, despite its excellent capability to reason about sound-emitting objects through the memory-augmented GDN, MAGNet achieves only modest performance gains while exhibiting the highest DSR when acoustically similar distractors are present.

\noindent\textbf{Comparison with Oracle settings.}
To further contextualize performance, we compare MAGNet with its oracle variants.
As shown in \cref{tab:clean_distracted}, Oracle2 achieves a substantial improvement over Oracle1, indicating that the limited duration of the goal sound significantly constrains navigation performance.
This finding underscores the importance of accurately updating goal information once the goal sound ceases.
Although MAGNet effectively maintains goal estimates after silence, a performance gap remains relative to Oracle1 and further widens against Oracle2, suggesting considerable room for improvement in the proposed GDN.

\subsection{Visualization of Navigation Trajectories}
\label{sec:visualization_of_navigation_trajectories}
We visualize the navigation trajectories of different methods under \emph{Clean Environments} in~\cref{fig:trajectories_clean}.
Intuitively, before the goal starts emitting sound, agents may either approach or diverge from it.
In the latter case (first row of \cref{fig:trajectories_clean}), additional actions are required within the limited sound-emitting window to compensate for earlier incorrect movements, leading to inefficient exploration. 
As shown in the second row of \cref{fig:trajectories_clean}, agents tend to lose their way once the goal sound ceases, ultimately failing to reach the goal.
In contrast, our method successfully completes the navigation task by leveraging historical context and self-motion cues to reliably maintain and update goal information.
See Supp. for more visualizations and video demos.

\begin{table*}[t]
    \centering
    \belowrulesep=0pt
    \aboverulesep=0pt
    \caption{Comparison of GDN performance for SAVi and MAGNet on the test split.
    \emph{Sounding} and \emph{Silent} denote metrics measured during the sound-emitting and silent periods, respectively.
    Note that a low $LE_{CD}$ accompanied by an extremely low $F_{\le 20^\circ}$ is not informative.}
    \resizebox{\textwidth}{!}{
    \begin{tabular}{c|c|ccccc|ccccc}
    \toprule
    \multirow{2}{*}{Period}
    & \multirow{2}{*}{Method}
    & \multicolumn{5}{c|}{\emph{Clean Environments}}
    & \multicolumn{5}{c}{\emph{Distracted Environments}} \\
    \cmidrule(lr){3-7} \cmidrule(lr){8-12}
    &
    & $ER_{\le 20^\circ}\raisebox{0.25ex}{$\downarrow$} $ 
    & $F_{\le 20^\circ}\raisebox{0.25ex}{$\uparrow$} $ 
    & $LE_{CD}\raisebox{0.25ex}{$\downarrow$} $ 
    & $LR_{CD}\raisebox{0.25ex}{$\uparrow$} $ 
    & $RDE\raisebox{0.25ex}{$\downarrow$} $ 

    & $ER_{\le 20^\circ}\raisebox{0.25ex}{$\downarrow$} $ 
    & $F_{\le 20^\circ}\raisebox{0.25ex}{$\uparrow$} $ 
    & $LE_{CD}\raisebox{0.25ex}{$\downarrow$} $ 
    & $LR_{CD}\raisebox{0.25ex}{$\uparrow$} $ 
    & $RDE\raisebox{0.25ex}{$\downarrow$} $  \\
    \midrule
    \multirow{2}{*}{Sounding}
    & SAVi
    & 0.882\hspace{1.0em} & 0.148\hspace{0.5em} & 39.71\hspace{0.5em} & 0.499\hspace{0.5em} & 0.270\hspace{0.5em} 
    & 1.123\hspace{1.0em} & 0.080\hspace{0.5em} & \textbf{43.12}\hspace{0.5em} & 0.515\hspace{0.5em} & \textbf{0.281}\hspace{0.5em} \\
    & MAGNet
    & \textbf{0.762}\hspace{1.0em} & \textbf{0.290}\hspace{0.5em} & \textbf{36.77}\hspace{0.5em} & \textbf{0.601}\hspace{0.5em} & \textbf{0.117}\hspace{0.5em} 
    & \textbf{0.896}\hspace{1.0em} & \textbf{0.119}\hspace{0.5em} & 47.84\hspace{0.5em} & \textbf{0.733}\hspace{0.5em} & 0.319\hspace{0.5em} \\
    \midrule
    \multirow{2}{*}{Silent}
    & SAVi
    & 0.995\hspace{1.0em} & 0.006\hspace{0.5em} & \textbf{18.05}\hspace{0.5em} & 0.015\hspace{0.5em} & 0.242\hspace{0.5em} 
    & 1.761\hspace{1.0em} & 0.027\hspace{0.5em} & 55.58\hspace{0.5em} & 0.240\hspace{0.5em} & \textbf{0.166}\hspace{0.5em} \\
    & MAGNet
    & \textbf{0.905}\hspace{1.0em} & \textbf{0.140}\hspace{0.5em} & 48.83\hspace{0.5em} & \textbf{0.368}\hspace{0.5em} & \textbf{0.166}\hspace{0.5em} 
    & \textbf{0.892}\hspace{1.0em} & \textbf{0.143}\hspace{0.5em} & \textbf{50.65}\hspace{0.5em} & \textbf{0.508}\hspace{0.5em} & 0.207\hspace{0.5em} \\
    \bottomrule
    \end{tabular}}

    \label{tab:goal_descriptor}
\end{table*}

\subsection{Factors Affecting Navigation Success}
\label{sec:factors_affecting_navigation_success}

To gain deeper insight into the factors that determine navigation success in SAVN-CE, we analyze two key variables that strongly influence SR: 
1) \textbf{Action Ratio}: the ratio between the oracle number of actions required to reach the goal and the number of actions available during the goal's sound-emitting period; 
2) \textbf{Geodesic Distance}: the geodesic distance between the agent's initial position and the goal. 
A higher action ratio indicates a larger portion of actions must be executed without the goal sound, making the episode inherently more challenging.
Similarly, a larger geodesic distance increases task difficulty, as sound attenuation and complex spatial layouts hinder accurate goal inference.

\begin{figure}[t]
    \centering
    \includegraphics[width=0.49\linewidth]{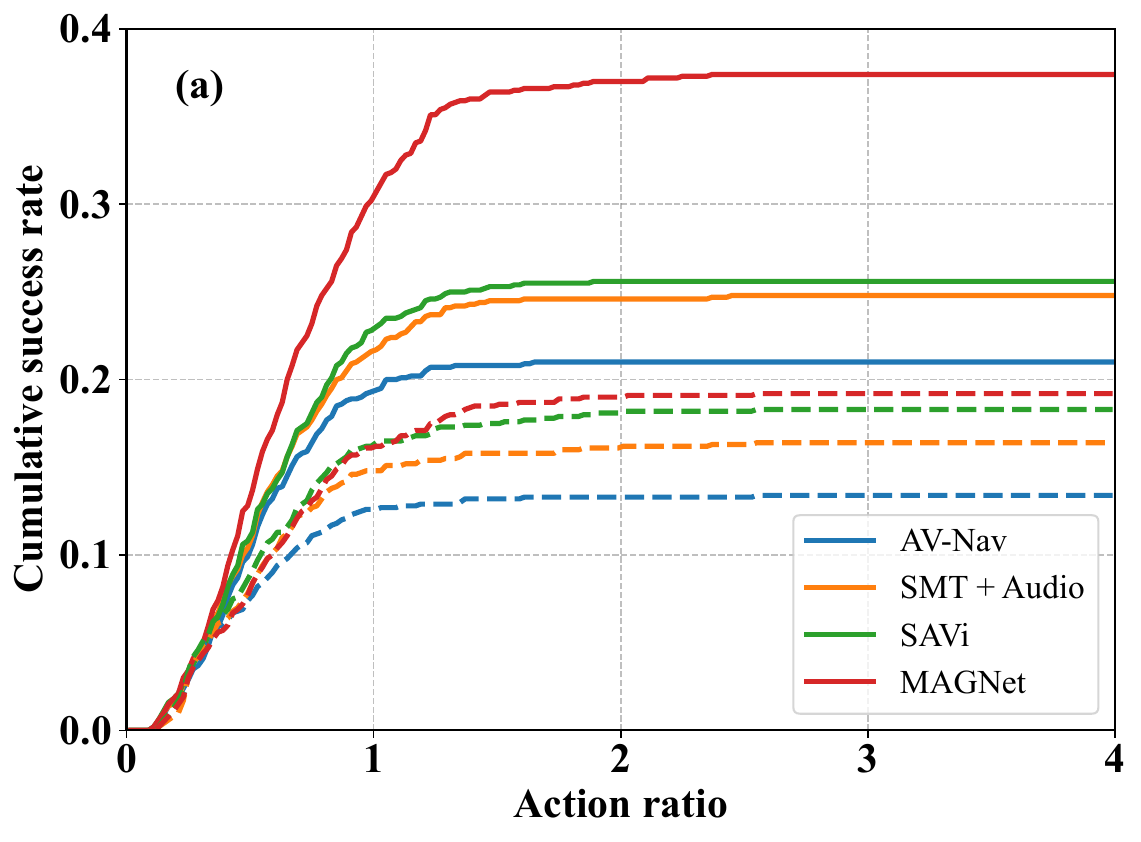}
    \includegraphics[width=0.49\linewidth]{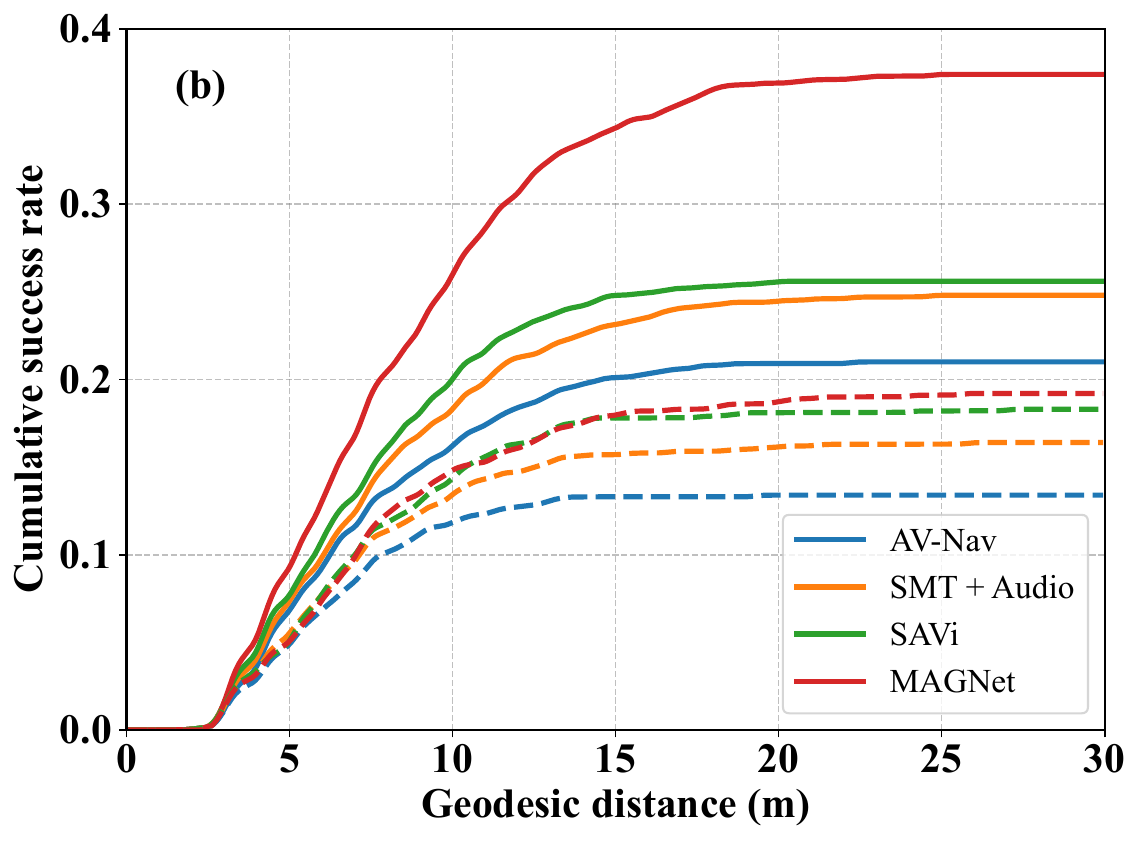}
    \caption{Impact of (a) action ratio and (b) geodesic distance on the cumulative success rates of different methods under \emph{Clean Environments} (solid) and \emph{Distracted Environments} (dashed).}
    \label{fig:cumulative_success_rate}
\end{figure}

\begin{table*}[t]
    \centering
    \belowrulesep=0pt
    \aboverulesep=0pt
    \caption{Ablation study of MAGNet on the SAVN-CE dataset under \emph{Clean Environments}.}
    \resizebox{\textwidth}{!}{
    \begin{tabular}{
        >{\centering\arraybackslash}p{0.6cm}
        >{\centering\arraybackslash}p{1.0cm}
        >{\centering\arraybackslash}p{1.8cm}
        |rrrrr|rrrrr}
    \toprule
    \multirow{2}{*}{GDN} & \multirow{2}{*}{Memory}  & \multirow{2}{*}{Self-motion} 
    & \multicolumn{5}{c|}{\emph{Navigation Metrics}} & \multicolumn{5}{c}{\emph{SELD Metrics}} \\
    \cmidrule(lr){4-8} \cmidrule(lr){9-13}
    & &
    & SR\raisebox{0.25ex}{$\uparrow$} 
    & SPL\raisebox{0.25ex}{$\uparrow$} 
    & SNA\raisebox{0.25ex}{$\uparrow$}                            
    & DTG\raisebox{0.25ex}{$\downarrow$} 
    & SWS\raisebox{0.25ex}{$\uparrow$}

    & $ER_{\le 20^\circ}\raisebox{0.25ex}{$\downarrow$} $ 
    & $F_{\le 20^\circ}\raisebox{0.25ex}{$\uparrow$} $ 
    & $LE_{CD}\raisebox{0.25ex}{$\downarrow$} $ 
    & $LR_{CD}\raisebox{0.25ex}{$\uparrow$} $ 
    & $RDE\raisebox{0.25ex}{$\downarrow$} $ \\
    \midrule
    \ding{55} & \ding{55}   & \ding{55}     
    & 32.4   & 27.9\hspace{0.4em}    & 23.5\hspace{0.5em}    & 9.2\hspace{0.5em}     & 6.3\hspace{0.5em} 
    & -\hspace{2.0em} & -\hspace{1.5em} & -\hspace{1.5em} & -\hspace{1.5em} & -\hspace{1.5em} \\

    \ding{51} & \ding{55}   & \ding{51}     
    & 33.9       & 29.8\hspace{0.4em}     & 24.8\hspace{0.5em}    & 8.8\hspace{0.5em}     & 8.9\hspace{0.5em}
    & 0.953\hspace{1.0em} & 0.082\hspace{0.5em} & \textbf{26.59}\hspace{0.5em} & 0.122\hspace{0.5em} & 0.128\hspace{0.5em} \\

    \ding{51} & \ding{51}   & \ding{55}     
    & 34.3      & 30.4\hspace{0.4em}     & 25.6\hspace{0.5em}    & 8.5\hspace{0.5em}     & 7.8\hspace{0.5em}
    & 0.855\hspace{1.0em} & 0.193\hspace{0.5em} & 36.15\hspace{0.5em} & 0.453\hspace{0.5em} & 0.134\hspace{0.5em} \\

    \ding{51} & \ding{51}   & \ding{51}     
    & \textbf{37.7} & \textbf{32.9}\hspace{0.4em} & \textbf{27.4}\hspace{0.5em} & \textbf{8.0}\hspace{0.5em} & \textbf{10.6}\hspace{0.5em}
    & \textbf{0.816}\hspace{1.0em} & \textbf{0.240}\hspace{0.5em} & 31.30\hspace{0.5em} & \textbf{0.494}\hspace{0.5em} & \textbf{0.119}\hspace{0.5em} \\

    \bottomrule
    \end{tabular}}
    \label{tab:ablation}
\end{table*}

The results presented in \cref{fig:cumulative_success_rate} reveal consistent trends across all evaluated methods. As either the action ratio or the geodesic distance increases, the cumulative success rate gradually saturates, indicating that most successful episodes occur with longer sound durations and closer goal distances. 
Compared with the baselines, MAGNet consistently achieves higher upper bounds on the cumulative success rate, demonstrating superior robustness to short-duration sounds and long-distance navigation scenarios. 
This highlights the capability of the memory-augmented GDN to maintain stable goal representations.

\subsection{Evaluation of the Goal Descriptor Network}
\label{sec:goal_descriptor_analysis}

Following standard SELD metrics~\cite{politis2021overview, krause2024sound}, we jointly assess the localization and detection performance of GDNs using the error rate and F1-score within a 20$^\circ$ localization tolerance ($ER_{\le 20^\circ}$ and $F_{\le 20^\circ}$), as well as the localization error, localization recall, and relative distance error conditioned on correctly detected events ($LE_{CD}$, $LR_{CD}$, and $RDE$).
The results, macro-averaged across goal categories, are summarized in \cref{tab:goal_descriptor}.  
See Supp. for a detailed analysis of the results and visualizations.

As shown in \cref{tab:goal_descriptor}, SELD metrics notably decline under \emph{Distracted Environments} compared with \emph{Clean Environments}, indicating that distractor sounds substantially increase the difficulty of accurate goal localization and categorization.
Our model outperforms SAVi on most metrics, demonstrating its superior capability to jointly infer spatial and semantic goal information.  
When the goal sound ceases, both methods experience a performance degradation; however, our memory-augmented GDN remains competitive, suggesting that it effectively preserves and updates goal representations even without auditory input.
These observations reinforce the motivation presented in \cref{sec:goal_descriptor_network}.

\subsection{Ablation Study}
\label{sec:ablation_study}

To further quantify the contribution of each component in MAGNet, 
we conduct a series of ablation experiments.  
The results summarized in \cref{tab:ablation} report both navigation and SELD performance under four configurations: 1) without the GDN, 2) with the GDN but no memory ($N_{g} \! = \! 1$), 3) with the GDN but without self-motion cues, and 4) the full memory-augmented GDN.  

Removing the GDN leads to a notable degradation in navigation performance; however, the remaining modules still outperform all baselines, indicating their effectiveness in capturing informative features for navigation.
Introducing either episodic memory or self-motion cues individually improves both navigation and SELD performance, highlighting their respective roles in modeling temporal context and spatial reasoning.  
Further improvements are achieved when all components are combined, demonstrating their complementary contributions within the memory-augmented GDN for robust navigation and SELD.
We also observe that improved SELD performance consistently translates into better navigation outcomes.
\section{Conclusion}
\label{sec:conclusion}
In this paper, we introduce SAVN-CE, a new task that extends semantic audio-visual navigation to continuous environments, where agents navigate freely toward semantically grounded sound-emitting goals while avoiding distractors.  
To address the challenge of losing goal information when the goal becomes silent, we propose MAGNet, which employs a memory-augmented goal descriptor network for robust goal reasoning and efficient navigation.  
Comprehensive experiments demonstrate that MAGNet substantially outperforms existing state-of-the-art methods, exhibiting strong robustness to short-duration sounds and long-distance navigation.  
Ablation studies further validate the effectiveness of our proposed method.
In future work, we plan to extend this framework to more complex auditory scenarios involving multiple or dynamic goals.
\clearpage
\section{Acknowledgements}
\label{sec:acknowledgements}

This work was supported in part by the Zhongguancun Academy (Grant No. 20240306),
the National Natural Science Foundation of China (Grant Nos. 62471340, 62376266, 62376140, U23A20315),
the Special Fund for Taishan Scholar Project of Shandong Province,
and the Tsinghua University-Toyota Research Center.
{
    \small
    \bibliographystyle{ieeenat_fullname}
    \bibliography{main.bbl}

\begin{thebibliography}{58}
\providecommand{\natexlab}[1]{#1}
\providecommand{\url}[1]{\texttt{#1}}
\expandafter\ifx\csname urlstyle\endcsname\relax
  \providecommand{\doi}[1]{doi: #1}\else
  \providecommand{\doi}{doi: \begingroup \urlstyle{rm}\Url}\fi

\bibitem[Adavanne et~al.(2018)Adavanne, Politis, and Virtanen]{adavanne2018direction}
Sharath Adavanne, Archontis Politis, and Tuomas Virtanen.
\newblock Direction of arrival estimation for multiple sound sources using convolutional recurrent neural network.
\newblock In \emph{Proc. EUSIPCO}, pages 1462--1466, 2018.

\bibitem[Adavanne et~al.(2019)Adavanne, Politis, Nikunen, and Virtanen]{adavanne2019sound}
Sharath Adavanne, Archontis Politis, Joonas Nikunen, and Tuomas Virtanen.
\newblock Sound event localization and detection of overlapping sources using convolutional recurrent neural networks.
\newblock \emph{IEEE J. Sel. Topics Signal Process.}, 13\penalty0 (1):\penalty0 34--48, 2019.

\bibitem[An et~al.(2024)An, Wang, Wang, Wang, Huang, He, and Wang]{an2024etpnav}
Dong An, Hanqing Wang, Wenguan Wang, Zun Wang, Yan Huang, Keji He, and Liang Wang.
\newblock {ETPNav}: {E}volving topological planning for vision-language navigation in continuous environments.
\newblock \emph{IEEE Trans. Pattern Anal. Mach. Intell.}, pages 5130--5145, 2024.

\bibitem[Anderson et~al.(2018{\natexlab{a}})Anderson, Chang, Chaplot, Dosovitskiy, Gupta, Koltun, Kosecka, Malik, Mottaghi, Savva, et~al.]{anderson2018evaluation}
Peter Anderson, Angel Chang, Devendra~Singh Chaplot, Alexey Dosovitskiy, Saurabh Gupta, Vladlen Koltun, Jana Kosecka, Jitendra Malik, Roozbeh Mottaghi, Manolis Savva, et~al.
\newblock On evaluation of embodied navigation agents.
\newblock \emph{arXiv preprint arXiv:1807.06757}, 2018{\natexlab{a}}.

\bibitem[Anderson et~al.(2018{\natexlab{b}})Anderson, Wu, Teney, Bruce, Johnson, S{\"u}nderhauf, Reid, Gould, and Van Den~Hengel]{anderson2018vision}
Peter Anderson, Qi Wu, Damien Teney, Jake Bruce, Mark Johnson, Niko S{\"u}nderhauf, Ian Reid, Stephen Gould, and Anton Van Den~Hengel.
\newblock Vision-and-language navigation: {Interpreting} visually-grounded navigation instructions in real environments.
\newblock In \emph{Proc. IEEE Conf. Comput. Vis. Pattern Recog.}, pages 3674--3683, 2018{\natexlab{b}}.

\bibitem[Berghi et~al.(2024)Berghi, Wu, Zhao, Wang, and Jackson]{berghi2024fusion}
Davide Berghi, Peipei Wu, Jinzheng Zhao, Wenwu Wang, and Philip~JB Jackson.
\newblock Fusion of audio and visual embeddings for sound event localization and detection.
\newblock In \emph{Proc. IEEE ICASSP}, pages 8816--8820, 2024.

\bibitem[Cakir et~al.(2017)Cakir, Parascandolo, Heittola, Huttunen, and Virtanen]{cakir2017convolutional}
Emre Cakir, Giambattista Parascandolo, Toni Heittola, Heikki Huttunen, and Tuomas Virtanen.
\newblock Convolutional recurrent neural networks for polyphonic sound event detection.
\newblock \emph{IEEE/ACM Trans. Audio, Speech, Lang. Process.}, 25\penalty0 (6):\penalty0 1291--1303, 2017.

\bibitem[Chang et~al.(2017)Chang, Dai, Funkhouser, Halber, Niebner, Savva, Song, Zeng, and Zhang]{chang2017matterport3d}
Angel Chang, Angela Dai, Thomas Funkhouser, Maciej Halber, Matthias Niebner, Manolis Savva, Shuran Song, Andy Zeng, and Yinda Zhang.
\newblock {Matterport3D}: {Learning} from {RGB-D} data in indoor environments.
\newblock In \emph{Proc. Int. Conf. 3D Vis.}, pages 667--676, 2017.

\bibitem[Chen et~al.(2020)Chen, Jain, Schissler, Gari, Al-Halah, Ithapu, Robinson, and Grauman]{chen2020soundspaces}
Changan Chen, Unnat Jain, Carl Schissler, Sebastia Vicenc~Amengual Gari, Ziad Al-Halah, Vamsi~Krishna Ithapu, Philip Robinson, and Kristen Grauman.
\newblock {SoundSpaces}: {A}udio-visual navigation in {3D} environments.
\newblock In \emph{Proc. Eur. Conf. Comput. Vis.}, pages 17--36, 2020.

\bibitem[Chen et~al.(2021{\natexlab{a}})Chen, Al-Halah, and Grauman]{chen2021semantic}
Changan Chen, Ziad Al-Halah, and Kristen Grauman.
\newblock Semantic audio-visual navigation.
\newblock In \emph{Proc. IEEE Conf. Comput. Vis. Pattern Recog.}, pages 15516--15525, 2021{\natexlab{a}}.

\bibitem[Chen et~al.(2021{\natexlab{b}})Chen, Majumder, Al-Halah, Gao, Ramakrishnan, and Grauman]{chen2021learning}
Changan Chen, Sagnik Majumder, Ziad Al-Halah, Ruohan Gao, Santhosh~Kumar Ramakrishnan, and Kristen Grauman.
\newblock Learning to set waypoints for audio-visual navigation.
\newblock In \emph{Proc. Int. Conf. Learn. Represent.}, pages 4861--4876, 2021{\natexlab{b}}.

\bibitem[Chen et~al.(2022{\natexlab{a}})Chen, Schissler, Garg, Kobernik, Clegg, Calamia, Batra, Robinson, and Grauman]{chen2022soundspaces}
Changan Chen, Carl Schissler, Sanchit Garg, Philip Kobernik, Alexander Clegg, Paul Calamia, Dhruv Batra, Philip Robinson, and Kristen Grauman.
\newblock {SoundSpaces 2.0}: {A} simulation platform for visual-acoustic learning.
\newblock In \emph{Proc. Adv. Neural Inform. Process. Syst.}, pages 8896--8911, 2022{\natexlab{a}}.

\bibitem[Chen et~al.(2024)Chen, Ramos, Tomar, and Grauman]{chen2024sim2real}
Changan Chen, Jordi Ramos, Anshul Tomar, and Kristen Grauman.
\newblock Sim2real transfer for audio-visual navigation with frequency-adaptive acoustic field prediction.
\newblock In \emph{Proc. IEEE/RSJ Int. Conf. Intell. Robots Syst.}, pages 8595--8602, 2024.

\bibitem[Chen et~al.(2025)Chen, An, Huang, Xu, Su, Ling, Reid, and Wang]{chen2025constraint}
Kehan Chen, Dong An, Yan Huang, Rongtao Xu, Yifei Su, Yonggen Ling, Ian Reid, and Liang Wang.
\newblock Constraint-aware zero-shot vision-language navigation in continuous environments.
\newblock \emph{IEEE Trans. Pattern Anal. Mach. Intell.}, 47\penalty0 (11):\penalty0 10441--10456, 2025.

\bibitem[Chen et~al.(2021{\natexlab{c}})Chen, Guhur, Schmid, and Laptev]{chen2021history}
Shizhe Chen, Pierre-Louis Guhur, Cordelia Schmid, and Ivan Laptev.
\newblock History aware multimodal transformer for vision-and-language navigation.
\newblock In \emph{Proc. Adv. Neural Inform. Process. Syst.}, pages 5834--5847, 2021{\natexlab{c}}.

\bibitem[Chen et~al.(2022{\natexlab{b}})Chen, Guhur, Tapaswi, Schmid, and Laptev]{chen2022think}
Shizhe Chen, Pierre-Louis Guhur, Makarand Tapaswi, Cordelia Schmid, and Ivan Laptev.
\newblock Think global, act local: {Dual}-scale graph transformer for vision-and-language navigation.
\newblock In \emph{Proc. IEEE Conf. Comput. Vis. Pattern Recog.}, pages 16537--16547, 2022{\natexlab{b}}.

\bibitem[Diaz-Guerra et~al.(2021)Diaz-Guerra, Miguel, and Beltran]{diaz2021gpurir}
David Diaz-Guerra, Antonio Miguel, and Jose~R Beltran.
\newblock {gpuRIR}: {A} python library for room impulse response simulation with {GPU} acceleration.
\newblock \emph{Multimedia Tools Applic.}, 80\penalty0 (4):\penalty0 5653--5671, 2021.

\bibitem[Fang et~al.(2019)Fang, Toshev, Fei-Fei, and Savarese]{fang2019scene}
Kuan Fang, Alexander Toshev, Li Fei-Fei, and Silvio Savarese.
\newblock Scene memory transformer for embodied agents in long-horizon tasks.
\newblock In \emph{Proc. IEEE Conf. Comput. Vis. Pattern Recog.}, pages 538--547, 2019.

\bibitem[Fried et~al.(2018)Fried, Hu, Cirik, Rohrbach, Andreas, Morency, Berg-Kirkpatrick, Saenko, Klein, and Darrell]{fried2018speaker}
Daniel Fried, Ronghang Hu, Volkan Cirik, Anna Rohrbach, Jacob Andreas, Louis-Philippe Morency, Taylor Berg-Kirkpatrick, Kate Saenko, Dan Klein, and Trevor Darrell.
\newblock Speaker-follower models for vision-and-language navigation.
\newblock In \emph{Proc. Adv. Neural Inform. Process. Syst.}, pages 3314--3325, 2018.

\bibitem[Gan et~al.(2020)Gan, Zhang, Wu, Gong, and Tenenbaum]{gan2020look}
Chuang Gan, Yiwei Zhang, Jiajun Wu, Boqing Gong, and Joshua~B Tenenbaum.
\newblock Look, listen, and act: {Towards} audio-visual embodied navigation.
\newblock In \emph{Proc. IEEE Int. Conf. Robotics Autom.}, pages 9701--9707, 2020.

\bibitem[Garc{\'\i}a-Barrios et~al.(2022)Garc{\'\i}a-Barrios, Krause, Politis, Mesaros, Guti{\'e}rrez-Arriola, and Fraile]{garcia2022binaural}
Guillermo Garc{\'\i}a-Barrios, Daniel~Aleksander Krause, Archontis Politis, Annamaria Mesaros, Juana~M Guti{\'e}rrez-Arriola, and Rub{\'e}n Fraile.
\newblock Binaural source localization using deep learning and head rotation information.
\newblock In \emph{Proc. EUSIPCO}, pages 36--40, 2022.

\bibitem[Georgakis et~al.(2022)Georgakis, Schmeckpeper, Wanchoo, Dan, Miltsakaki, Roth, and Daniilidis]{georgakis2022cross}
Georgios Georgakis, Karl Schmeckpeper, Karan Wanchoo, Soham Dan, Eleni Miltsakaki, Dan Roth, and Kostas Daniilidis.
\newblock Cross-modal map learning for vision and language navigation.
\newblock In \emph{Proc. IEEE Conf. Comput. Vis. Pattern Recog.}, pages 15460--15470, 2022.

\bibitem[Grumiaux et~al.(2022)Grumiaux, Kiti{\'c}, Girin, and Gu{\'e}rin]{grumiaux2022survey}
Pierre-Amaury Grumiaux, Sr{\dj}an Kiti{\'c}, Laurent Girin, and Alexandre Gu{\'e}rin.
\newblock A survey of sound source localization with deep learning methods.
\newblock \emph{J. Acoust. Soc. Am.}, 152\penalty0 (1):\penalty0 107--151, 2022.

\bibitem[Gupta et~al.(2017)Gupta, Davidson, Levine, Sukthankar, and Malik]{gupta2017cognitive}
Saurabh Gupta, James Davidson, Sergey Levine, Rahul Sukthankar, and Jitendra Malik.
\newblock Cognitive mapping and planning for visual navigation.
\newblock In \emph{Proc. IEEE Conf. Comput. Vis. Pattern Recog.}, pages 2616--2625, 2017.

\bibitem[He et~al.(2016)He, Zhang, Ren, and Sun]{he2016deep}
Kaiming He, Xiangyu Zhang, Shaoqing Ren, and Jian Sun.
\newblock Deep residual learning for image recognition.
\newblock In \emph{Proc. IEEE Conf. Comput. Vis. Pattern Recog.}, pages 770--778, 2016.

\bibitem[Hong et~al.(2021)Hong, Wu, Qi, Rodriguez-Opazo, and Gould]{hong2021vln}
Yicong Hong, Qi Wu, Yuankai Qi, Cristian Rodriguez-Opazo, and Stephen Gould.
\newblock {VLN-BERT}: {A} recurrent vision-and-language bert for navigation.
\newblock In \emph{Proc. IEEE Conf. Comput. Vis. Pattern Recog.}, pages 1643--1653, 2021.

\bibitem[Hong et~al.(2022)Hong, Wang, Wu, and Gould]{hong2022bridging}
Yicong Hong, Zun Wang, Qi Wu, and Stephen Gould.
\newblock Bridging the gap between learning in discrete and continuous environments for vision-and-language navigation.
\newblock In \emph{Proc. IEEE Conf. Comput. Vis. Pattern Recog.}, pages 15439--15449, 2022.

\bibitem[Hu et~al.(2025)Hu, Cao, Wu, Kang, Yang, Wang, Plumbley, and Yang]{hu2025pseldnets}
Jinbo Hu, Yin Cao, Ming Wu, Fang Kang, Feiran Yang, Wenwu Wang, Mark~D Plumbley, and Jun Yang.
\newblock {PSELDNets}: {P}re-trained neural networks on a large-scale synthetic dataset for sound event localization and detection.
\newblock \emph{IEEE Trans. Audio, Speech, Lang. Process.}, 33:\penalty0 2845--2860, 2025.

\bibitem[Kingma and Ba(2015)]{kingma2015adam}
Diederik~P. Kingma and Jimmy~Lei Ba.
\newblock Adam: {A} method for stochastic optimization.
\newblock In \emph{Proc. Int. Conf. Learn. Represent.}, pages 1--15, 2015.

\bibitem[Kondoh and Kanezaki(2023)]{kondoh2023multi}
Haru Kondoh and Asako Kanezaki.
\newblock Multi-goal audio-visual navigation using sound direction map.
\newblock In \emph{Proc. IEEE/RSJ Int. Conf. Intell. Robots Syst.}, pages 5219--5226, 2023.

\bibitem[Krantz and Lee(2022)]{krantz2022sim}
Jacob Krantz and Stefan Lee.
\newblock {Sim-2-Sim} transfer for vision-and-language navigation in continuous environments.
\newblock In \emph{Proc. Eur. Conf. Comput. Vis.}, pages 588--603, 2022.

\bibitem[Krantz et~al.(2020)Krantz, Wijmans, Majumdar, Batra, and Lee]{krantz2020beyond}
Jacob Krantz, Erik Wijmans, Arjun Majumdar, Dhruv Batra, and Stefan Lee.
\newblock Beyond the {Nav-Graph}: {V}ision-and-language navigation in continuous environments.
\newblock In \emph{Proc. Eur. Conf. Comput. Vis.}, pages 104--120, 2020.

\bibitem[Krantz et~al.(2021)Krantz, Gokaslan, Batra, Lee, and Maksymets]{krantz2021waypoint}
Jacob Krantz, Aaron Gokaslan, Dhruv Batra, Stefan Lee, and Oleksandr Maksymets.
\newblock Waypoint models for instruction-guided navigation in continuous environments.
\newblock In \emph{Proc. Int. Conf. Comput. Vis.}, pages 15162--15171, 2021.

\bibitem[Krause et~al.(2023)Krause, Garc{\'\i}a-Barrios, Politis, and Mesaros]{krause2023binaural}
Daniel~Aleksander Krause, Guillermo Garc{\'\i}a-Barrios, Archontis Politis, and Annamaria Mesaros.
\newblock Binaural sound source distance estimation and localization for a moving listener.
\newblock \emph{IEEE/ACM Trans. Audio, Speech, Lang. Process.}, 32:\penalty0 996--1011, 2023.

\bibitem[Krause et~al.(2024)Krause, Politis, and Mesaros]{krause2024sound}
Daniel~Aleksander Krause, Archontis Politis, and Annamaria Mesaros.
\newblock Sound event detection and localization with distance estimation.
\newblock In \emph{Proc. EUSIPCO}, pages 286--290, 2024.

\bibitem[Liu et~al.(2024)Liu, Paul, Chatterjee, and Cherian]{liu2024caven}
Xiulong Liu, Sudipta Paul, Moitreya Chatterjee, and Anoop Cherian.
\newblock {CAVEN}: {An} embodied conversational agent for efficient audio-visual navigation in noisy environments.
\newblock In \emph{Proc. AAAI}, pages 3765--3773, 2024.

\bibitem[Pashevich et~al.(2021)Pashevich, Schmid, and Sun]{pashevich2021episodic}
Alexander Pashevich, Cordelia Schmid, and Chen Sun.
\newblock Episodic transformer for vision-and-language navigation.
\newblock In \emph{Proc. Int. Conf. Comput. Vis.}, pages 15942--15952, 2021.

\bibitem[Paul et~al.(2022)Paul, Roy-Chowdhury, and Cherian]{paul2022avlen}
Sudipta Paul, Amit Roy-Chowdhury, and Anoop Cherian.
\newblock {AVLEN}: {Audio-visual-language} embodied navigation in {3D} environments.
\newblock In \emph{Proc. Adv. Neural Inform. Process. Syst.}, pages 6236--6249, 2022.

\bibitem[Perotin et~al.(2019)Perotin, Serizel, Vincent, and Guerin]{perotin2019crnn}
Laureline Perotin, Romain Serizel, Emmanuel Vincent, and Alexandre Guerin.
\newblock {CRNN}-based multiple {DOA} estimation using acoustic intensity features for {Ambisonics} recordings.
\newblock \emph{IEEE J. Sel. Topics Signal Process.}, 13\penalty0 (1):\penalty0 22--33, 2019.

\bibitem[Politis et~al.(2021)Politis, Mesaros, Adavanne, Heittola, and Virtanen]{politis2021overview}
Archontis Politis, Annamaria Mesaros, Sharath Adavanne, Toni Heittola, and Tuomas Virtanen.
\newblock Overview and evaluation of sound event localization and detection in {DCASE} 2019.
\newblock \emph{IEEE/ACM Trans. Audio, Speech, Lang. Process.}, 29:\penalty0 684--698, 2021.

\bibitem[Savva et~al.(2019)Savva, Kadian, Maksymets, Zhao, Wijmans, Jain, Straub, Liu, Koltun, Malik, et~al.]{savva2019habitat}
Manolis Savva, Abhishek Kadian, Oleksandr Maksymets, Yili Zhao, Erik Wijmans, Bhavana Jain, Julian Straub, Jia Liu, Vladlen Koltun, Jitendra Malik, et~al.
\newblock Habitat: {A} platform for embodied {AI} research.
\newblock In \emph{Proc. Int. Conf. Comput. Vis.}, pages 9339--9347, 2019.

\bibitem[Scheibler et~al.(2018)Scheibler, Bezzam, and Dokmani{\'c}]{scheibler2018pyroomacoustics}
Robin Scheibler, Eric Bezzam, and Ivan Dokmani{\'c}.
\newblock Pyroomacoustics: {A} python package for audio room simulation and array processing algorithms.
\newblock In \emph{Proc. IEEE ICASSP}, pages 351--355, 2018.

\bibitem[Schulman et~al.(2017)Schulman, Wolski, Dhariwal, Radford, and Klimov]{schulman2017proximal}
John Schulman, Filip Wolski, Prafulla Dhariwal, Alec Radford, and Oleg Klimov.
\newblock Proximal policy optimization algorithms.
\newblock \emph{arXiv preprint arXiv:1707.06347}, 2017.

\bibitem[Shi et~al.(2025)Shi, Zhang, Li, and Shen]{shi2025towards}
Zhanbo Shi, Lin Zhang, Linfei Li, and Ying Shen.
\newblock Towards audio-visual navigation in noisy environments: {A} large-scale benchmark dataset and an architecture considering multiple sound-sources.
\newblock In \emph{Proc. AAAI}, pages 14673--14680, 2025.

\bibitem[Shimada et~al.(2021)Shimada, Koyama, Takahashi, Takahashi, and Mitsufuji]{shimada2021accdoa}
Kazuki Shimada, Yuichiro Koyama, Naoya Takahashi, Shusuke Takahashi, and Yuki Mitsufuji.
\newblock {ACCDOA}: {Activity}-coupled {Cartesian} direction of arrival representation for sound event localization and detection.
\newblock In \emph{Proc. IEEE ICASSP}, pages 915--919, 2021.

\bibitem[Shimada et~al.(2022)Shimada, Koyama, Takahashi, Takahashi, Tsunoo, and Mitsufuji]{shimada2022multi}
Kazuki Shimada, Yuichiro Koyama, Shusuke Takahashi, Naoya Takahashi, Emiru Tsunoo, and Yuki Mitsufuji.
\newblock Multi-{ACCDOA}: Localizing and detecting overlapping sounds from the same class with auxiliary duplicating permutation invariant training.
\newblock In \emph{Proc. IEEE ICASSP}, pages 316--320, 2022.

\bibitem[Shimada et~al.(2023)Shimada, Politis, Sudarsanam, Krause, Uchida, Adavanne, Hakala, Koyama, Takahashi, Takahashi, Virtanen, and Mitsufuji]{shimada2023starss23}
Kazuki Shimada, Archontis Politis, Parthasaarathy Sudarsanam, Daniel~A. Krause, Kengo Uchida, Sharath Adavanne, Aapo Hakala, Yuichiro Koyama, Naoya Takahashi, Shusuke Takahashi, Tuomas Virtanen, and Yuki Mitsufuji.
\newblock {STARSS23}: {An} audio-visual dataset of spatial recordings of real scenes with spatiotemporal annotations of sound events.
\newblock In \emph{Proc. Adv. Neural Inform. Process. Syst.}, pages 72931--72957, 2023.

\bibitem[Szot et~al.(2021)Szot, Clegg, Undersander, Wijmans, Zhao, Turner, Maestre, Mukadam, Chaplot, Maksymets, et~al.]{szot2021habitat}
Andrew Szot, Alexander Clegg, Eric Undersander, Erik Wijmans, Yili Zhao, John Turner, Noah Maestre, Mustafa Mukadam, Devendra~Singh Chaplot, Oleksandr Maksymets, et~al.
\newblock Habitat 2.0: {T}raining home assistants to rearrange their habitat.
\newblock In \emph{Proc. Adv. Neural Inform. Process. Syst.}, pages 251--266, 2021.

\bibitem[Wang et~al.(2019)Wang, Huang, Celikyilmaz, Gao, Shen, Wang, Wang, and Zhang]{wang2019reinforced}
Xin Wang, Qiuyuan Huang, Asli Celikyilmaz, Jianfeng Gao, Dinghan Shen, Yuan-Fang Wang, William~Yang Wang, and Lei Zhang.
\newblock Reinforced cross-modal matching and self-supervised imitation learning for vision-language navigation.
\newblock In \emph{Proc. IEEE Conf. Comput. Vis. Pattern Recog.}, pages 6629--6638, 2019.

\bibitem[Wijmans et~al.(2020)Wijmans, Kadian, Morcos, Lee, Essa, Parikh, Savva, and Batra]{wijmans2020ddppo}
Erik Wijmans, Abhishek Kadian, Ari Morcos, Stefan Lee, Irfan Essa, Devi Parikh, Manolis Savva, and Dhruv Batra.
\newblock {DD-PPO}: {Learning} near-perfect pointgoal navigators from 2.5 billion frames.
\newblock In \emph{Proc. Int. Conf. Learn. Represent.}, pages 5648--5668, 2020.

\bibitem[Wu et~al.(2019)Wu, Wu, Tamar, Russell, Gkioxari, and Tian]{wu2019bayesian}
Yi Wu, Yuxin Wu, Aviv Tamar, Stuart Russell, Georgia Gkioxari, and Yuandong Tian.
\newblock Bayesian relational memory for semantic visual navigation.
\newblock In \emph{Proc. Int. Conf. Comput. Vis.}, pages 2769--2779, 2019.

\bibitem[Xia et~al.(2020)Xia, Togneri, Sohel, Zhao, and Huang]{xia2020sound}
Xianjun Xia, Roberto Togneri, Ferdous Sohel, Yuanjun Zhao, and Defeng Huang.
\newblock Sound event detection using multiple optimized kernels.
\newblock \emph{IEEE/ACM Trans. Audio, Speech, Lang. Process.}, 28:\penalty0 1745--1754, 2020.

\bibitem[Yang et~al.(2024)Yang, Liu, Chen, Cherian, Marks, Le~Roux, and Gan]{yang2024rila}
Zeyuan Yang, Jiageng Liu, Peihao Chen, Anoop Cherian, Tim~K Marks, Jonathan Le~Roux, and Chuang Gan.
\newblock {RILA}: {Reflective} and imaginative language agent for zero-shot semantic audio-visual navigation.
\newblock In \emph{Proc. IEEE Conf. Comput. Vis. Pattern Recog.}, pages 16251--16261, 2024.

\bibitem[Younes et~al.(2023)Younes, Honerkamp, Welschehold, and Valada]{younes2023catch}
Abdelrahman Younes, Daniel Honerkamp, Tim Welschehold, and Abhinav Valada.
\newblock Catch me if you hear me: {Audio-visual} navigation in complex unmapped environments with moving sounds.
\newblock \emph{IEEE Robot. Autom. Lett.}, 8\penalty0 (2):\penalty0 928--935, 2023.

\bibitem[Yu et~al.(2022)Yu, Huang, Sun, Chen, Wang, and Liu]{yu2022sound}
Yinfeng Yu, Wenbing Huang, Fuchun Sun, Changan Chen, Yikai Wang, and Xiaohong Liu.
\newblock Sound adversarial audio-visual navigation.
\newblock In \emph{Proc. Int. Conf. Learn. Represent.}, pages 26129--26152, 2022.

\bibitem[Yue et~al.(2024)Yue, Zhou, Xie, Zhang, Yan, and Yin]{yue2024safe}
Lu Yue, Dongliang Zhou, Liang Xie, Feitian Zhang, Ye Yan, and Erwei Yin.
\newblock Safe-{VLN}: {Collision} avoidance for vision-and-language navigation of autonomous robots operating in continuous environments.
\newblock \emph{IEEE Robot. Autom. Lett.}, 9\penalty0 (6):\penalty0 4918--4925, 2024.

\bibitem[Zhu et~al.(2017{\natexlab{a}})Zhu, Gordon, Kolve, Fox, Fei-Fei, Gupta, Mottaghi, and Farhadi]{zhu2017visual}
Yuke Zhu, Daniel Gordon, Eric Kolve, Dieter Fox, Li Fei-Fei, Abhinav Gupta, Roozbeh Mottaghi, and Ali Farhadi.
\newblock Visual semantic planning using deep successor representations.
\newblock In \emph{Proc. Int. Conf. Comput. Vis.}, pages 483--492, 2017{\natexlab{a}}.

\bibitem[Zhu et~al.(2017{\natexlab{b}})Zhu, Mottaghi, Kolve, Lim, Gupta, Fei-Fei, and Farhadi]{zhu2017target}
Yuke Zhu, Roozbeh Mottaghi, Eric Kolve, Joseph~J Lim, Abhinav Gupta, Li Fei-Fei, and Ali Farhadi.
\newblock Target-driven visual navigation in indoor scenes using deep reinforcement learning.
\newblock In \emph{Proc. IEEE Int. Conf. Robotics Autom.}, pages 3357--3364, 2017{\natexlab{b}}.

\end{thebibliography}
}

\end{document}